\documentclass{article}
\usepackage{stywhispers}
\usepackage{amsmath}
\usepackage{epsfig}
\usepackage{amssymb}
\usepackage{booktabs}
\usepackage{multirow}
\usepackage{graphicx}
\usepackage{rotating}
\usepackage{subcaption}
\usepackage{stfloats}
\usepackage{float}
\usepackage{graphicx}
\usepackage{url}
\usepackage{xcolor}
\usepackage[table]{xcolor}
\usepackage{graphicx}
\usepackage{multirow}
\usepackage{array}
\usepackage[table]{xcolor}
\usepackage{pgf}
\usepackage{hyperref}
\usepackage{xcolor}
\usepackage{soul}
\usepackage{cite}

\definecolor{heatred}{RGB}{178,24,43}
\definecolor{heatblue}{RGB}{33,102,172}

\newcommand{\heatcell}[2]{%
  \pgfmathtruncatemacro{\heatpct}{round(min(55,55*abs(#1)/3))}%
  \ifdim #1pt<0pt
    \edef\heatshade{heatred!\heatpct!white}%
    \expandafter\cellcolor\expandafter{\heatshade}#2%
  \else
    \ifdim #1pt>0pt
      \edef\heatshade{heatblue!\heatpct!white}%
      \expandafter\cellcolor\expandafter{\heatshade}#2%
    \else
      \cellcolor{white}#2%
    \fi
  \fi
}

\title{HSI-Road Relabeled: Surface-Aware Road-Scene Segmentation}
\name{Imad Ali Shah~\textsuperscript{1,*},
Imran Mehmood~\textsuperscript{1},
Enda Ward~\textsuperscript{2},
Martin Glavin~\textsuperscript{1},
Edward Jones~\textsuperscript{1} and
Brian Deegan~\textsuperscript{1}
\thanks{
\newline
This work was funded, in part, by Taighde Éireann--Research Ireland grants 13/RC/2094\_P2 and 18/SP/5942, and Valeo Vision Systems.
\newline Annotations, data splits, and registration metadata are made available at~\scriptsize\url{https://github.com/imadalishah/HSI_Road_relabeled}.
\newline Correspondence: i.shah2@universityofgalway.ie
}
}
\address{\textsuperscript{1} School of Engineering and Ryan Institute, University of Galway, Ireland\\
\textsuperscript{2} Valeo Vision Systems, Tuam, Ireland}

\begin{document}
\maketitle
\begin{abstract}
The HSI-Road dataset provides paired RGB and 25-channel NIR (680--960~nm) images with binary masks but no surface-level labels.~This paper introduces a manually labeled six-class taxonomy: Background, Asphalt, Concrete, Dirt, Water, and Grass, and an RGB-to-NIR registration pipeline with corresponding annotations. Six semantic-segmentation models are evaluated under four input configurations: original-resolution RGB, registered low-resolution RGB (RGB$_{\text{reg}}$), NIR, and channel-stacked RGB$_{\text{reg}}$--NIR (RGBN$_{\text{stk}}$).~The comparison quantifies the effect of spatial-resolution reduction on RGB, along with evaluation of NIR and RGBN$_{\text{stk}}$, with results reported using per-class and mean IoU and F1 scores.~The original-resolution RGB achieves the highest overall performance but contains 12$\times$ more pixels and incurs a 15.5--20.6$\%$ latency penalty compared to the reduced-resolution inputs.~At the common 192$\times$384 resolution, RGBN$_{\text{stk}}$ outperforms NIR for all six models and RGB$_{\text{reg}}$ for five of six, with the most consistent gains for the Water class. These results highlight the importance of spatial resolution while showing that NIR provides complementary information to RGB.

\end{abstract}
\vspace{-1ex} 

\vspace{-1.25ex}
\section{Introduction}
\vspace{-1ex}
\label{sec:intro}
Road-surface understanding is an important aspect of autonomous driving (AD), especially for traction control and road-condition monitoring. The public hyperspectral imaging (HSI) driving datasets~\cite{shah2025hyperspectral} generally represent the road as a single semantic class rather than distinguishing road-surface materials. This supports free-space detection but not surface-aware tasks such as distinguishing paved from unpaved terrain or detecting standing water and drivable grassy patches. Spectral imaging such as near-infrared (NIR) may complement RGB due to different materials' reflectance across visible and NIR wavelengths, helping distinguish surfaces that may appear similar in RGB. However, their robustness requires empirical evaluation for each class and downstream architecture.

HSI-Road~\cite{lu2020hsi} is one of the largest public HSI datasets for AD~\cite{shah2024hyperspectral} and contains various drivable surfaces (such as asphalt, dirt, etc), but its binary masks do not support material-level segmentation.
In addition, HSI-Road has not previously been extended with multiclass surface-oriented annotations. This gap motivates the current work, with its main contributions as follows:


\vspace{-1.25ex}
\begin{enumerate}
    \item Six-class,~surface-oriented HSI-Road annotations, along with an RGB-to-NIR registration pipeline.
    \vspace{-1.35ex}
    \item Four-input configuration-based evaluation: (i) RGB$_{\text{ori}}$: Original dataset resolution RGB, (ii) RGB$_{\text{reg}}$: Registered low-resolution RGB, (iii) NIR; and (iv) RGBN$_{\text{stk}}$: Channel-stacked RGB$_{\text{reg}}$--NIR.
    \vspace{-1.35ex}
    \item Multilabel-stratified train/validation/test split that approximately preserves image-level class prevalence.
\end{enumerate}

\vspace{-3ex}
\section{Background and Motivation}
\vspace{-1ex}
\label{sec:format}
\subsection{RGB and NIR for Road Scene Understanding}
\vspace{-1ex}

Remote sensing widely combines NIR and RGB modalities to discriminate materials\cite{valme2025adas} such as vegetation, moisture, and mineral surfaces, leveraging their distinct spectral characteristics. However, its exploration in the AD domain remains limited~\cite{shah2025hyperspectral}.~Prior works using available HSI datasets~\cite{shah2024hyperspectral} for road scenes generally target drivable areas (e.g., roads) versus other broader multiclass urban-object segmentation. Moreover, explicit matched-resolution comparisons between RGB$_{\text{reg}}$ and NIR for surface-oriented multiclass segmentation remain underexplored. This gap makes it difficult to separate pure multimodal gains from performance degradation caused by registration artifacts and spatial-resolution mismatch. Recent studies demonstrate that spatial misregistration between modalities can severely degrade detection performance~\cite{usama2026misalignment}. Consequently, explicit RGB--NIR alignment is important for pixel-level fusion approaches.

\vspace{-1.25ex}
\subsection{HSI-Road Dataset and HSI Segmentation in AD}
\vspace{-1ex}
HSI-Road contains 3799 paired RGB ($704\times1280$) and 25-channel NIR ($192\times384$) images with pixel-level binary masks: Background (Bkg.) and Road. It covers asphalt, concrete (Conc.), and dirt track, with over 60\% images on rural roads~\cite{lu2020hsi}. Despite this diversity, the binary labelling scheme restricts the dataset to drivable-area segmentation, motivating extension to road surface-oriented annotations.

Unlike extensive exploration of deep learning (DL) in other domains~\cite{valme2025adas}, DL in HSI for AD remains relatively limited~\cite{shah2025hyperspectral}. Previous studies have considered pixel-level ANN-based classification and baseline semantic segmentation models (SSMs)~\cite{shah2024hyperspectral}, including UNet~\cite{ronneberger2015u}, DeepLabV3+~\cite{chen2017deeplab}, 
and SegFormer~\cite{xie2021segformer}. These efforts demonstrate increasing interest in HSI-based road-scene semantic understanding.

\begin{table}[htbp]
\centering
\caption{Pixel-level class distribution comparison between the original binary and the relabeled multi-class annotations.}
\vspace{-1ex}
\label{tab:dataset_class_distribution}
\resizebox{\columnwidth}{!}{
\small
\begin{tabular}{p{3.1ex} p{16.9ex} rr rr}
\toprule
\multirow{2}{*}{\textbf{Label}} & \multirow{2}{*}{\textbf{  Class}} & \multicolumn{2}{c}{\textbf{NIR ($192 \times 384$)}} & \multicolumn{2}{c}{\textbf{RGB ($704 \times 1280$)}} \\
\cmidrule(lr){3-4} \cmidrule(lr){5-6}
& & \textbf{Pixels} & \textbf{\%} & \textbf{Pixels} & \textbf{\%} \\
\midrule
\multicolumn{6}{l}{\textit{Original Binary Class Distribution}} \\
0 & Bkg. & 204,185,446 & 72.90 & 2,513,487,955 & 73.42 \\
1 & Road & 75,907,226 & 27.10 & 909,866,925 & 26.58 \\
\midrule
\multicolumn{6}{l}{\textit{Relabeled Multi-Class Distribution}} \\
0 & Bkg.~\textsuperscript{1} & 201,076,168 & 71.79 & 2,474,507,082 & 72.28 \\
1 & Asphalt   & 27,689,446  & 9.89  & 331,948,381   & 9.70  \\
2 & Conc.~\textsuperscript{2}  & 15,547,453  & 5.55  & 188,276,659   & 5.50  \\
3 & Dirt~\textsuperscript{3}  & 33,587,271  & 11.99 & 402,433,271   & 11.76 \\
4 & Water~\textsuperscript{4} & 794,868     & 0.28  & 9,553,815     & 0.28  \\
5 & Grass~\textsuperscript{5} & 1,397,466   & 0.50  & 16,635,672    & 0.49  \\
\bottomrule
\end{tabular}
}
\parbox{\columnwidth}{\raggedright \footnotesize
\textsuperscript{1} Include all pixels not assigned to the other five target surface classes.\\
\textsuperscript{2} Concrete. \quad
\textsuperscript{3} Unpaved dirt, soil, and sand surfaces. \quad
\textsuperscript{4} Standing water and roadside drains/ditches. \quad
\textsuperscript{5} Roadside grassy regions considered drivable under the adopted annotation taxonomy.
}
\end{table}

\vspace{-4ex}
\section{Annotation and Modality Alignment}
\vspace{-1ex}
\subsection{Six-Class Surface-Oriented Taxonomy}
\vspace{-1ex}
\label{sec:taxonomy}
The relabeled annotation contains six mutually exclusive pixel classes: Bkg., Asphalt, Conc., Dirt, Water, and Grass. Table~\ref{tab:dataset_class_distribution} compares the original binary and the relabeled multiclass annotations, and reports the corresponding pixel distributions. The six-class annotations were created manually using RGB$_{\text{ori}}$, which provides finer spatial detail than NIR, with a custom segmentation tool.
The relabeled annotations are not a strict subdivision of the original Road class, and include redrawn surface boundaries under the new annotation taxonomy. Consequently, some pixels originally labeled Bkg. are assigned to surface classes, such as Water and Grass. The relabeling, therefore, modifies both semantic granularity and the local spatial extent of the annotated surface region.

\vspace{-1.25ex}
\subsection{Spatial Registration: RGB-to-NIR Modality}
\vspace{-1ex}
\label{sec:registration}
RGB was co-registered to the NIR resolution.~A pseudo-NIR grayscale was formed by averaging three channels, initially using 0, 14, and 23~\footnote{Combination selected empirically after testing 100 random images.}. However, if post-registration Normalized Cross-Correlation (NCC) was $<0.70$, remaining triplets among the $\binom{25}{3}-1=2299$ combinations were tested until NCC$\geq0.70$. If no triplet reached this threshold, the highest-NCC based triplet was retained~\footnote{Metadata in the codebase includes image-wise best triplets.}.

\begin{figure}[htbp]
    \centering
    \begin{subfigure}[b]{0.495\columnwidth}
        \centering
        \includegraphics[width=\textwidth]{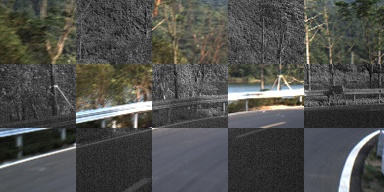} 
        \label{fig:sub_a}
    \end{subfigure}
    \begin{subfigure}[b]{0.495\columnwidth}
        \centering
        \includegraphics[width=\textwidth]{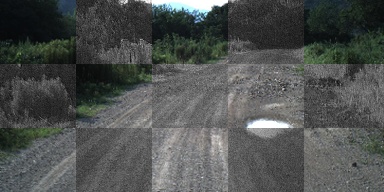} 
        \label{fig:sub_b}
    \end{subfigure}
    \vspace{-6ex}
    \caption{Checkerboard visualization of spatial alignment using alternating RGB and NIR image patches.}
    \label{fig:checkerboard_visualization}
\end{figure}

Contrast-enhanced grayscales were matched using Oriented FAST and Rotated BRIEF (ORB) with Scale-Invariant Feature Transform (SIFT) fallback, and Random Sample Consensus (RANSAC) estimated the initial affine transform under conservative geometric constraints.~Enhanced Correlation Coefficient (ECC) refinement using Euclidean and translation models estimated a residual transform that was composed with the feature-based transform.~Across all mapped RGB--NIR pairs, the mean~$\pm$~SD NCC was $0.73\pm0.12$. The mean inlier reprojection residual was $1.22\pm0.15$~px on the NIR grid. Example alignments are shown in Fig.~\ref{fig:checkerboard_visualization}. The same mapping was applied to semantic masks using nearest-neighbor interpolation to preserve categorical labels.

\vspace{-2ex}
\begin{table}[ht]
\centering
\caption{Image-level class distribution across dataset splits.}
\vspace{-1ex}
\label{tab:stratified_class_distribution}
\resizebox{\columnwidth}{!}{
\begin{tabular}{l cc cc cc}
\toprule
\multirow{2}{*}{\textbf{Class}} & \multicolumn{2}{c}{\textbf{Train} ($N=2659: 70\%$)} & \multicolumn{2}{c}{\textbf{Val} ($N=570: 15\%$)} & \multicolumn{2}{c}{\textbf{Test} ($N=570: 15\%$)} \\
\cmidrule(lr){2-3} \cmidrule(lr){4-5} \cmidrule(lr){6-7}
 & \textbf{Count} & \textbf{\%} & \textbf{Count} & \textbf{\%} & \textbf{Count} & \textbf{\%} \\
\midrule
Bkg. & 2659 & 100 & 570 & 100 & 570 & 100 \\
Asphalt      & 816  & 30.69  & 175 & 30.70  & 175 & 30.70  \\
Conc.  & 770  & 28.96  & 165 & 28.95  & 165 & 28.95  \\
Dirt      & 1578 & 59.35  & 321 & 56.32  & 323 & 56.67  \\
Water     & 727  & 27.34  & 156 & 27.37  & 156 & 27.37  \\
Grass     & 484  & 18.20  & 103 & 18.07  & 104 & 18.25  \\
\bottomrule
\end{tabular}
}
\end{table}

\vspace{-2ex}
\begin{table}[htbp]
\centering
\caption{Training hyperparameters and input configurations.}
\vspace{-1ex}
\label{tab:training_hyperparameters}
\resizebox{\columnwidth}{!}{
\begin{tabular}{ll}
\toprule
\textbf{Hyperparameter} & \textbf{Value / Range} \\
\midrule
\multicolumn{2}{l}{\textit{Optimization \& Training}} \\
Epochs / Early Stop & Max 300 / 50 epochs on validation mIoU \\
Batch Size / Seeds   & 16 / 3 (42, 43, 44) \\
Optimizer / Rate  & AdamW / $1 \times 10^{-4}$ \\
Loss Function   & Avg of Cross-Entropy and Dice Loss \\
Data split & Stratified 70/15/15 (see in Table~\ref{tab:stratified_class_distribution}) \\
Precision & Automatic mixed precision \\
\midrule
\multicolumn{2}{l}{\textit{SSMs Input ($C, H, W$)}} \\
RGB$_{\text{ori}}$              & (3, 704, 1280) \\
RGB$_{\text{reg}}$ and NIR     & (3, 192, 384) and (25, 192, 384) \\
RGBN$_{\text{stk}}$              & (28, 192, 384) \\
Horizontal/Vertical Flip & $p = 0.5$ (at dataloader) \\
\midrule
\multicolumn{2}{l}{\textit{Hardware \& Environment}} \\
GPU & NVIDIA RTX PRO 6000 Blackwell Max-Q Edition \\
CPU & Dual Intel Xeon Gold 6252 (48C/96T, 2.10 GHz) \\
OS/workers & Debian 12 and 4 Data Loader Workers \\
\bottomrule
\end{tabular}
}
\end{table}

\vspace{-3ex}
\section{Experimental Setup and Results}
\vspace{-1ex}
\subsection{Segmentation Backbones}
\vspace{-1ex}
\label{sec:backbone}
Six SSMs are evaluated on the relabeled annotations of the HSI-Road dataset: UNet, UNet-CBAM~\cite{woo2018cbam,shah2024hyperspectral}, DeepLabV3+, SegFormer, and two UPerNet~\cite{xiao2018unified} variants using EfficientNet-B0 and MiT-B0 encoders. UNet, UNet-CBAM, and DeepLabV3+ use EfficientNet-B0, whereas SegFormer uses MiT-B0. To evaluate the effect of encoder choice while holding the decoder fixed, UPerNet is evaluated with EfficientNet-B0 (UPerNet$_{\text{EN-B0}}$) as the CNN and MiT-B0 (UPerNet$_{\text{MiT-B0}}$) as the transformer encoder. For NIR and RGBN$_{\text{stk}}$, the first layer is adapted to accept 25 and 28 channels, respectively.




\vspace{-1.25ex}
\subsection{Data Split, Model Training and Evaluation}
\vspace{-1ex}
\label{sec:training}
The 3799 images are divided into train/validation/test subsets containing 2659/570/570 images, respectively, as shown in Table~\ref{tab:stratified_class_distribution}, using multilabel stratification based on image-level class presence. Since the image-wise split is not provided in the original dataset, the codebase of this work includes it.

All SSMs use the training settings reported in Table~\ref{tab:training_hyperparameters} and are trained three times from scratch using different seeds. All six SSMs are evaluated using four input configurations: (1)~RGB$_{\text{ori}}$; (2)~RGB$_{\text{reg}}$; (3)~NIR; and (4)~RGBN$_{\text{stk}}$. The RGB$_{\text{ori}}$--to--RGB$_{\text{reg}}$ comparison reflects the combined effects of spatial resizing, registration, interpolation, residual misregistration, and transformation of the corresponding relabeled masks. RGB$_{\text{reg}}$, NIR, and RGBN$_{\text{stk}}$ use the same $192\times384$ spatial resolution and therefore provide matched resolution-based input modality comparisons.  In addition to evaluation on relabeled masks, the experiments also include original binary labels on RGB$_{\text{ori}}$, NIR, and RGBN$_{\text{stk}}$ input modalities.



\vspace{-1ex}
\begin{table}[htbp]
\caption{Modality-wise evaluation on the binary taxonomy. mIoU/mF1: dataset-pooled\textsuperscript{*} macro mean$\pm$SD across seeds.}
\vspace{-1ex}
\centering
\resizebox{\linewidth}{!}{%
\begin{tabular}{p{1cm}p{1.91cm}cccccc}
\toprule
\multirow{2}{*}{\textbf{Modality}} & \multirow{2}{*}{\textbf{ SSM}} &
\multirow{2}{*}{\textbf{mIoU}} & \multirow{2}{*}{\textbf{mF1}} &
\multicolumn{2}{c}{\textbf{IoU}} & \multicolumn{2}{c}{\textbf{F1}} \\
\cmidrule(lr){5-6} \cmidrule(lr){7-8}
 & & & & \textbf{Bkg.} & \textbf{Road} & \textbf{Bkg.} & \textbf{Road} \\
\midrule

\multirow{6}{*}{RGB$_{\text{ori}}$}
 & UNet
 & 96.86{\scriptsize$\pm$0.04}
 & 98.40{\scriptsize$\pm$0.02}
 & 98.32{\scriptsize$\pm$0.02}
 & 95.41{\scriptsize$\pm$0.05}
 & 99.15{\scriptsize$\pm$0.01}
 & 97.65{\scriptsize$\pm$0.03} \\

 & UNet-CBAM
 & 96.77{\scriptsize$\pm$0.06}
 & 98.36{\scriptsize$\pm$0.03}
 & 98.28{\scriptsize$\pm$0.03}
 & 95.27{\scriptsize$\pm$0.09}
 & 99.13{\scriptsize$\pm$0.02}
 & 97.58{\scriptsize$\pm$0.05} \\

 & DeepLabV3+
 & \textbf{96.94{\scriptsize$\pm$0.02}}
 & \textbf{98.44{\scriptsize$\pm$0.01}}
 & \textbf{98.37{\scriptsize$\pm$0.02}}
 & \textbf{95.51{\scriptsize$\pm$0.03}}
 & \textbf{99.18{\scriptsize$\pm$0.01}}
 & \textbf{97.70{\scriptsize$\pm$0.02}} \\

 & SegFormer
 & 96.76{\scriptsize$\pm$0.03}
 & 98.35{\scriptsize$\pm$0.01}
 & 98.26{\scriptsize$\pm$0.01}
 & 95.25{\scriptsize$\pm$0.04}
 & 99.12{\scriptsize$\pm$0.01}
 & 97.57{\scriptsize$\pm$0.02} \\

 & UPerNet$_{\text{EN-B0}}$
 & 96.89{\scriptsize$\pm$0.05}
 & 98.42{\scriptsize$\pm$0.02}
 & 98.34{\scriptsize$\pm$0.03}
 & 95.44{\scriptsize$\pm$0.07}
 & 99.16{\scriptsize$\pm$0.01}
 & 97.67{\scriptsize$\pm$0.03} \\

 & UPerNet$_{\text{MiT-B0}}$
 & 96.90{\scriptsize$\pm$0.05}
 & 98.42{\scriptsize$\pm$0.03}
 & 98.34{\scriptsize$\pm$0.03}
 & 95.45{\scriptsize$\pm$0.07}
 & 99.16{\scriptsize$\pm$0.01}
 & 97.67{\scriptsize$\pm$0.04} \\

\midrule

\multirow{6}{*}{NIR}
 & UNet
 & 95.84{\scriptsize$\pm$0.11}
 & 97.87{\scriptsize$\pm$0.06}
 & 97.71{\scriptsize$\pm$0.06}
 & 93.97{\scriptsize$\pm$0.15}
 & 98.84{\scriptsize$\pm$0.03}
 & 96.89{\scriptsize$\pm$0.08} \\

 & UNet-CBAM
 & 95.81{\scriptsize$\pm$0.12}
 & 97.85{\scriptsize$\pm$0.06}
 & 97.69{\scriptsize$\pm$0.07}
 & 93.93{\scriptsize$\pm$0.17}
 & 98.83{\scriptsize$\pm$0.04}
 & 96.87{\scriptsize$\pm$0.09} \\

 & DeepLabV3+
 & 95.87{\scriptsize$\pm$0.14}
 & 97.88{\scriptsize$\pm$0.07}
 & 97.73{\scriptsize$\pm$0.07}
 & 94.01{\scriptsize$\pm$0.20}
 & 98.85{\scriptsize$\pm$0.04}
 & 96.91{\scriptsize$\pm$0.11} \\

 & SegFormer
 & 95.89{\scriptsize$\pm$0.03}
 & 97.89{\scriptsize$\pm$0.02}
 & 97.74{\scriptsize$\pm$0.02}
 & 94.04{\scriptsize$\pm$0.04}
 & 98.86{\scriptsize$\pm$0.01}
 & 96.93{\scriptsize$\pm$0.02} \\

 & UPerNet$_{\text{EN-B0}}$
 & 95.97{\scriptsize$\pm$0.02}
 & 97.93{\scriptsize$\pm$0.01}
 & 97.79{\scriptsize$\pm$0.01}
 & 94.15{\scriptsize$\pm$0.03}
 & 98.88{\scriptsize$\pm$0.00}
 & 96.99{\scriptsize$\pm$0.01} \\

 & UPerNet$_{\text{MiT-B0}}$
 & \textbf{96.02{\scriptsize$\pm$0.02}}
 & \textbf{97.96{\scriptsize$\pm$0.01}}
 & \textbf{97.82{\scriptsize$\pm$0.01}}
 & \textbf{94.22{\scriptsize$\pm$0.03}}
 & \textbf{98.90{\scriptsize$\pm$0.01}}
 & \textbf{97.03{\scriptsize$\pm$0.02}} \\

\midrule

\multirow{6}{*}{RGBN$_{\text{stk}}$}
 & UNet
 & 96.09{\scriptsize$\pm$0.02}
 & 98.00{\scriptsize$\pm$0.01}
 & 97.85{\scriptsize$\pm$0.01}
 & 94.32{\scriptsize$\pm$0.04}
 & 98.92{\scriptsize$\pm$0.00}
 & 97.08{\scriptsize$\pm$0.02} \\

 & UNet-CBAM
 & 95.95{\scriptsize$\pm$0.05}
 & 97.92{\scriptsize$\pm$0.03}
 & 97.78{\scriptsize$\pm$0.03}
 & 94.12{\scriptsize$\pm$0.07}
 & 98.88{\scriptsize$\pm$0.02}
 & 96.97{\scriptsize$\pm$0.04} \\

 & DeepLabV3+
 & 96.01{\scriptsize$\pm$0.09}
 & 97.95{\scriptsize$\pm$0.05}
 & 97.80{\scriptsize$\pm$0.05}
 & 94.21{\scriptsize$\pm$0.13}
 & 98.89{\scriptsize$\pm$0.03}
 & 97.02{\scriptsize$\pm$0.07} \\

 & SegFormer
 & 96.04{\scriptsize$\pm$0.05}
 & 97.97{\scriptsize$\pm$0.03}
 & 97.82{\scriptsize$\pm$0.03}
 & 94.25{\scriptsize$\pm$0.07}
 & 98.90{\scriptsize$\pm$0.02}
 & 97.04{\scriptsize$\pm$0.04} \\

 & UPerNet$_{\text{EN-B0}}$
 & 96.19{\scriptsize$\pm$0.04}
 & 98.05{\scriptsize$\pm$0.02}
 & 97.91{\scriptsize$\pm$0.02}
 & 94.46{\scriptsize$\pm$0.05}
 & 98.94{\scriptsize$\pm$0.01}
 & 97.15{\scriptsize$\pm$0.03} \\

 & UPerNet$_{\text{MiT-B0}}$ 
 & \textbf{96.21{\scriptsize$\pm$0.06}}
 & \textbf{98.06{\scriptsize$\pm$0.03}}
 & \textbf{97.92{\scriptsize$\pm$0.03}}
 & \textbf{94.49{\scriptsize$\pm$0.08}}
 & \textbf{98.95{\scriptsize$\pm$0.02}}
 & \textbf{97.17{\scriptsize$\pm$0.04}} \\

\bottomrule
\end{tabular}%
}


\parbox{\columnwidth}{\raggedright \footnotesize
\textsuperscript{*} For each seed, class-wise metrics are pooled over all pixels in the test set.
}
\label{tab:results_2class}
\end{table}

\vspace{-6ex}
\subsection{Overall Performance}
\vspace{-1ex}
Tables~\ref{tab:results_2class} and~\ref{tab:results_6class} report performance on the original binary and relabeled six-class masks, respectively. Binary mask performance is consistently high and tightly clustered, with 95.81--96.94\% mIoU and 97.85--98.44\% mF1 across all evaluated SSM-to-modality combinations, consistent with prior studies~\cite{shah2024hyperspectral}. For RGB$_{\text{ori}}$, DeepLabV3+ achieves the highest mIoU at 96.94\%, followed by UPerNet$_{\text{MiT-B0}}$ at 96.90\%. UPerNet$_{\text{MiT-B0}}$ also achieves the highest binary mIoU for NIR of 96.02\% and for RGBN$_{\text{stk}}$ of 96.21\%. However, the narrow spread of 1.13 mIoU and 0.59 mF1 points indicates that the results on binary masks provide limited performance insights across the evaluated SSM-to-modality combinations.



In contrast, the six-class taxonomy yields a wider performance range, with mIoU from 73.39--82.73\% and mF1 from 81.70--89.43\%. The remainder of the analysis therefore focuses on the six-class taxonomy only.~UPerNet$_{\text{MiT-B0}}$ achieves~the highest mIoU for three input configurations (RGB$_{\text{ori}}$: 82.73$\pm~$0.40\%, RGB$_{\text{reg}}$: 77.03$\pm$0.27\%, and RGBN$_{\text{stk}}$: 79.66$\pm$0.46\%), while UPerNet$_{\text{EN-B0}}$ performs best for NIR with 77.37$\pm$0.38\%. Across input modalities, Bkg., Asphalt, Conc., and Dirt retain relatively high IoU, whereas Water and Grass are substantially more difficult to segment. Although the best RGBN$_{\text{stk}}$ SSM remains 3.07 mIoU points below RGB$_{\text{ori}}$, the two operate at different spatial resolutions. At the matched $192\times384$ resolution, RGBN$_{\text{stk}}$ exceeds RGB$_{\text{reg}}$ by 2.63 mIoU points for UPerNet$_{\text{MiT-B0}}$, indicating that NIR complements the RGB modality and increases performance.


\begin{table*}[htbp]
\caption{Modality-wise evaluation on the six-class taxonomy. mIoU/mF1: dataset-pooled macro mean$\pm$SD across seeds.}
\vspace{-1ex}
\centering
\resizebox{\linewidth}{!}{%
\begin{tabular}{p{1cm}p{1.91cm}cccccccccccccc}
\toprule
\multirow{2}{*}{\textbf{Modality}} & \multirow{2}{*}{\textbf{ SSMs}} & \multirow{2}{*}{\textbf{mIoU}} & \multirow{2}{*}{\textbf{mF1}} & \multicolumn{6}{c}{\textbf{IoU}} & \multicolumn{6}{c}{\textbf{F1}} \\
\cmidrule(lr){5-10} \cmidrule(lr){11-16}
 & & & & \textbf{Bkg.} & \textbf{Asphalt} & \textbf{Conc.} & \textbf{Dirt} & \textbf{Water} & \textbf{Grass} & \textbf{Bkg.} & \textbf{Asphalt} & \textbf{Conc.} & \textbf{Dirt} & \textbf{Water} & \textbf{Grass} \\
\midrule

\multirow{6}{*}{RGB$_{\text{ori}}$}
 & UNet & 79.86{\scriptsize$\pm$0.23} & 87.19{\scriptsize$\pm$0.20} & 97.67{\scriptsize$\pm$0.01} & 96.10{\scriptsize$\pm$0.29} & 92.88{\scriptsize$\pm$0.51} & 90.23{\scriptsize$\pm$0.40} & 47.34{\scriptsize$\pm$0.39} & 54.94{\scriptsize$\pm$2.27} & 98.82{\scriptsize$\pm$0.00} & 98.01{\scriptsize$\pm$0.15} & 96.31{\scriptsize$\pm$0.27} & 94.87{\scriptsize$\pm$0.22} & 64.26{\scriptsize$\pm$0.35} & 70.90{\scriptsize$\pm$1.89} \\
 & UNet-CBAM & 79.78{\scriptsize$\pm$0.47} & 87.17{\scriptsize$\pm$0.47} & 97.67{\scriptsize$\pm$0.02} & 95.81{\scriptsize$\pm$0.49} & 92.59{\scriptsize$\pm$0.49} & 90.09{\scriptsize$\pm$0.48} & 48.03{\scriptsize$\pm$2.78} & 54.49{\scriptsize$\pm$1.48} & 98.82{\scriptsize$\pm$0.01} & 97.86{\scriptsize$\pm$0.26} & 96.15{\scriptsize$\pm$0.26} & 94.79{\scriptsize$\pm$0.27} & 64.86{\scriptsize$\pm$2.52} & 70.53{\scriptsize$\pm$1.25} \\
 & DeepLabV3+ & 81.33{\scriptsize$\pm$0.86} & 88.37{\scriptsize$\pm$0.62} & 97.83{\scriptsize$\pm$0.03} & 96.08{\scriptsize$\pm$0.74} & 93.02{\scriptsize$\pm$1.08} & 91.07{\scriptsize$\pm$0.21} & 49.81{\scriptsize$\pm$2.14} & 60.18{\scriptsize$\pm$2.75} & 98.91{\scriptsize$\pm$0.02} & 98.00{\scriptsize$\pm$0.38} & 96.38{\scriptsize$\pm$0.58} & 95.32{\scriptsize$\pm$0.11} & 66.48{\scriptsize$\pm$1.90} & 75.12{\scriptsize$\pm$2.16} \\
 & SegFormer & 79.37{\scriptsize$\pm$0.62} & 86.84{\scriptsize$\pm$0.54} & 97.61{\scriptsize$\pm$0.01} & 95.54{\scriptsize$\pm$0.54} & 91.73{\scriptsize$\pm$0.68} & 90.36{\scriptsize$\pm$0.31} & 46.02{\scriptsize$\pm$2.37} & 54.95{\scriptsize$\pm$1.40} & 98.79{\scriptsize$\pm$0.01} & 97.72{\scriptsize$\pm$0.28} & 95.69{\scriptsize$\pm$0.37} & 94.93{\scriptsize$\pm$0.17} & 63.00{\scriptsize$\pm$2.24} & 70.92{\scriptsize$\pm$1.17} \\
 & UPerNet$_{\text{EN-B0}}$ & 82.25{\scriptsize$\pm$0.55} & 89.02{\scriptsize$\pm$0.46} & 97.93{\scriptsize$\pm$0.01} & 96.55{\scriptsize$\pm$0.19} & \textbf{93.79{\scriptsize$\pm$0.34}} & \textbf{91.61{\scriptsize$\pm$0.04}} & 51.09{\scriptsize$\pm$2.49} & 62.50{\scriptsize$\pm$0.71} & 98.96{\scriptsize$\pm$0.00} & 98.24{\scriptsize$\pm$0.10} & \textbf{96.79{\scriptsize$\pm$0.18}} & \textbf{95.62{\scriptsize$\pm$0.02}} & 67.60{\scriptsize$\pm$2.20} & 76.92{\scriptsize$\pm$0.54} \\
 & UPerNet$_{\text{MiT-B0}}$ & \textbf{82.73{\scriptsize$\pm$0.40}} & \textbf{89.43{\scriptsize$\pm$0.33}} & \textbf{97.97{\scriptsize$\pm$0.03}} & \textbf{96.73{\scriptsize$\pm$0.18}} & 93.77{\scriptsize$\pm$0.33} & 91.51{\scriptsize$\pm$0.16} & \textbf{53.25{\scriptsize$\pm$1.25}} & \textbf{63.18{\scriptsize$\pm$1.80}} & \textbf{98.97{\scriptsize$\pm$0.01}} & \textbf{98.34{\scriptsize$\pm$0.10}} & 96.78{\scriptsize$\pm$0.17} & 95.57{\scriptsize$\pm$0.09} & \textbf{69.49{\scriptsize$\pm$1.06}} & \textbf{77.42{\scriptsize$\pm$1.36}} \\

\midrule
\multirow{6}{*}{RGB$_{\text{reg}}$}
 & UNet & 75.42{\scriptsize$\pm$1.01} & 83.49{\scriptsize$\pm$0.76} & 97.20{\scriptsize$\pm$0.05} & 93.49{\scriptsize$\pm$1.03} & 89.64{\scriptsize$\pm$1.58} & 88.61{\scriptsize$\pm$0.70} & 35.14{\scriptsize$\pm$2.88} & 48.47{\scriptsize$\pm$3.42} & 98.58{\scriptsize$\pm$0.03} & 96.63{\scriptsize$\pm$0.55} & 94.53{\scriptsize$\pm$0.88} & 93.96{\scriptsize$\pm$0.40} & 51.96{\scriptsize$\pm$3.16} & 65.25{\scriptsize$\pm$3.10} \\
 & UNet-CBAM & 75.77{\scriptsize$\pm$1.59} & 83.60{\scriptsize$\pm$1.56} & 97.16{\scriptsize$\pm$0.11} & 94.40{\scriptsize$\pm$0.52} & 91.02{\scriptsize$\pm$0.82} & 88.51{\scriptsize$\pm$0.64} & 33.84{\scriptsize$\pm$5.94} & 49.67{\scriptsize$\pm$1.87} & 98.56{\scriptsize$\pm$0.06} & 97.12{\scriptsize$\pm$0.27} & 95.30{\scriptsize$\pm$0.45} & 93.91{\scriptsize$\pm$0.36} & 50.37{\scriptsize$\pm$6.78} & 66.36{\scriptsize$\pm$1.67} \\
 & DeepLabV3+ & 76.61{\scriptsize$\pm$0.85} & 84.41{\scriptsize$\pm$0.75} & 97.22{\scriptsize$\pm$0.08} & \textbf{94.50{\scriptsize$\pm$0.60}} & \textbf{91.11{\scriptsize$\pm$0.78}} & \textbf{89.20{\scriptsize$\pm$0.25}} & 36.23{\scriptsize$\pm$2.60} & \textbf{51.39{\scriptsize$\pm$1.28}} & 98.59{\scriptsize$\pm$0.04} & \textbf{97.17{\scriptsize$\pm$0.32}} & \textbf{95.35{\scriptsize$\pm$0.43}} & \textbf{94.29{\scriptsize$\pm$0.14}} & 53.16{\scriptsize$\pm$2.82} & \textbf{67.89{\scriptsize$\pm$1.11}} \\
 & SegFormer & 74.70{\scriptsize$\pm$0.81} & 83.05{\scriptsize$\pm$0.72} & 97.18{\scriptsize$\pm$0.09} & 92.28{\scriptsize$\pm$0.60} & 87.70{\scriptsize$\pm$1.33} & 87.96{\scriptsize$\pm$0.36} & 34.96{\scriptsize$\pm$1.96} & 48.10{\scriptsize$\pm$1.91} & 98.57{\scriptsize$\pm$0.05} & 95.99{\scriptsize$\pm$0.32} & 93.45{\scriptsize$\pm$0.76} & 93.60{\scriptsize$\pm$0.21} & 51.79{\scriptsize$\pm$2.14} & 64.94{\scriptsize$\pm$1.73} \\
 & UPerNet$_{\text{EN-B0}}$ & 76.28{\scriptsize$\pm$1.06} & 84.22{\scriptsize$\pm$1.01} & 97.40{\scriptsize$\pm$0.05} & 94.16{\scriptsize$\pm$0.86} & 90.34{\scriptsize$\pm$0.53} & 88.88{\scriptsize$\pm$0.74} & 38.40{\scriptsize$\pm$4.68} & 48.54{\scriptsize$\pm$4.85} & 98.68{\scriptsize$\pm$0.02} & 96.99{\scriptsize$\pm$0.46} & 94.92{\scriptsize$\pm$0.29} & 94.11{\scriptsize$\pm$0.42} & 55.38{\scriptsize$\pm$4.97} & 65.26{\scriptsize$\pm$4.48} \\
 & UPerNet$_{\text{MiT-B0}}$ & \textbf{77.03{\scriptsize$\pm$0.27}} & \textbf{84.99{\scriptsize$\pm$0.24}} & \textbf{97.45{\scriptsize$\pm$0.06}} & 93.94{\scriptsize$\pm$0.23} & 90.25{\scriptsize$\pm$1.05} & 88.92{\scriptsize$\pm$0.35} & \textbf{40.87{\scriptsize$\pm$0.75}} & 50.74{\scriptsize$\pm$2.29} & \textbf{98.71{\scriptsize$\pm$0.03}} & 96.88{\scriptsize$\pm$0.12} & 94.88{\scriptsize$\pm$0.58} & 94.13{\scriptsize$\pm$0.19} & \textbf{58.02{\scriptsize$\pm$0.75}} & 67.30{\scriptsize$\pm$2.00} \\

\cmidrule(lr){2-16}
\multirow{6}{*}{NIR}
 & UNet & 73.39{\scriptsize$\pm$0.86} & 81.70{\scriptsize$\pm$0.92} & 96.66{\scriptsize$\pm$0.14} & 93.05{\scriptsize$\pm$0.41} & 88.35{\scriptsize$\pm$2.33} & 87.36{\scriptsize$\pm$0.58} & 31.82{\scriptsize$\pm$4.23} & 43.12{\scriptsize$\pm$1.56} & 98.30{\scriptsize$\pm$0.07} & 96.40{\scriptsize$\pm$0.22} & 93.80{\scriptsize$\pm$1.32} & 93.25{\scriptsize$\pm$0.33} & 48.17{\scriptsize$\pm$4.84} & 60.24{\scriptsize$\pm$1.53} \\
 & UNet-CBAM & 73.71{\scriptsize$\pm$1.42} & 82.02{\scriptsize$\pm$1.17} & 96.59{\scriptsize$\pm$0.18} & 92.48{\scriptsize$\pm$1.61} & 89.23{\scriptsize$\pm$1.04} & 87.10{\scriptsize$\pm$1.79} & 32.28{\scriptsize$\pm$2.46} & 44.55{\scriptsize$\pm$2.70} & 98.26{\scriptsize$\pm$0.09} & 96.09{\scriptsize$\pm$0.87} & 94.31{\scriptsize$\pm$0.58} & 93.10{\scriptsize$\pm$1.03} & 48.77{\scriptsize$\pm$2.79} & 61.61{\scriptsize$\pm$2.56} \\
 & DeepLabV3+ & 75.19{\scriptsize$\pm$0.42} & 83.31{\scriptsize$\pm$0.32} & 96.72{\scriptsize$\pm$0.09} & 93.22{\scriptsize$\pm$0.50} & \textbf{89.86{\scriptsize$\pm$1.45}} & 88.43{\scriptsize$\pm$0.42} & 34.79{\scriptsize$\pm$0.60} & 48.11{\scriptsize$\pm$3.19} & 98.33{\scriptsize$\pm$0.05} & 96.49{\scriptsize$\pm$0.27} & \textbf{94.65{\scriptsize$\pm$0.80}} & 93.86{\scriptsize$\pm$0.24} & 51.61{\scriptsize$\pm$0.66} & 64.92{\scriptsize$\pm$2.90} \\
 & SegFormer & 75.64{\scriptsize$\pm$1.44} & 83.85{\scriptsize$\pm$1.21} & 96.85{\scriptsize$\pm$0.17} & 93.45{\scriptsize$\pm$0.79} & 89.43{\scriptsize$\pm$1.49} & \textbf{88.49{\scriptsize$\pm$1.02}} & 40.98{\scriptsize$\pm$2.02} & 44.66{\scriptsize$\pm$4.15} & 98.40{\scriptsize$\pm$0.09} & 96.61{\scriptsize$\pm$0.42} & 94.42{\scriptsize$\pm$0.83} & \textbf{93.89{\scriptsize$\pm$0.58}} & 58.12{\scriptsize$\pm$2.03} & 61.67{\scriptsize$\pm$3.94} \\
 & UPerNet$_{\text{EN-B0}}$ & \textbf{77.37{\scriptsize$\pm$0.38}} & \textbf{85.51{\scriptsize$\pm$0.23}} & \textbf{96.99{\scriptsize$\pm$0.07}} & \textbf{93.48{\scriptsize$\pm$0.57}} & 88.94{\scriptsize$\pm$1.99} & 88.47{\scriptsize$\pm$0.50} & \textbf{45.64{\scriptsize$\pm$0.89}} & \textbf{50.69{\scriptsize$\pm$1.19}} & \textbf{98.47{\scriptsize$\pm$0.03}} & \textbf{96.63{\scriptsize$\pm$0.30}} & 94.14{\scriptsize$\pm$1.12} & 93.88{\scriptsize$\pm$0.28} & \textbf{62.67{\scriptsize$\pm$0.84}} & \textbf{67.27{\scriptsize$\pm$1.04}} \\
 & UPerNet$_{\text{MiT-B0}}$ & 73.42{\scriptsize$\pm$3.11} & 82.06{\scriptsize$\pm$2.97} & 96.50{\scriptsize$\pm$0.87} & 91.90{\scriptsize$\pm$0.94} & 86.46{\scriptsize$\pm$2.12} & 85.60{\scriptsize$\pm$4.17} & 36.61{\scriptsize$\pm$11.76} & 43.47{\scriptsize$\pm$2.53} & 98.22{\scriptsize$\pm$0.45} & 95.78{\scriptsize$\pm$0.51} & 92.73{\scriptsize$\pm$1.23} & 92.20{\scriptsize$\pm$2.44} & 52.85{\scriptsize$\pm$13.16} & 60.57{\scriptsize$\pm$2.46} \\

\cmidrule(lr){2-16}
\multirow{6}{*}{RGBN$_{\text{stk}}$}
 & UNet & 76.86{\scriptsize$\pm$0.66} & 84.70{\scriptsize$\pm$0.52} & 97.19{\scriptsize$\pm$0.06} & 94.92{\scriptsize$\pm$0.42} & 91.33{\scriptsize$\pm$0.79} & 89.20{\scriptsize$\pm$0.50} & 40.40{\scriptsize$\pm$2.38} & 48.13{\scriptsize$\pm$1.94} & 98.58{\scriptsize$\pm$0.03} & 97.39{\scriptsize$\pm$0.22} & 95.47{\scriptsize$\pm$0.43} & 94.29{\scriptsize$\pm$0.28} & 57.53{\scriptsize$\pm$2.40} & 64.96{\scriptsize$\pm$1.75} \\
 & UNet-CBAM & 75.63{\scriptsize$\pm$0.98} & 83.57{\scriptsize$\pm$0.87} & 97.12{\scriptsize$\pm$0.09} & 94.43{\scriptsize$\pm$0.95} & 90.85{\scriptsize$\pm$1.14} & 89.07{\scriptsize$\pm$0.13} & 37.86{\scriptsize$\pm$0.34} & 44.45{\scriptsize$\pm$5.22} & 98.54{\scriptsize$\pm$0.05} & 97.13{\scriptsize$\pm$0.50} & 95.20{\scriptsize$\pm$0.62} & 94.22{\scriptsize$\pm$0.07} & 54.93{\scriptsize$\pm$0.36} & 61.42{\scriptsize$\pm$4.94} \\
 & DeepLabV3+ & 77.85{\scriptsize$\pm$0.83} & 85.51{\scriptsize$\pm$0.78} & 97.29{\scriptsize$\pm$0.05} & 95.00{\scriptsize$\pm$0.06} & \textbf{91.81{\scriptsize$\pm$0.58}} & 89.78{\scriptsize$\pm$0.17} & 40.65{\scriptsize$\pm$3.51} & 52.60{\scriptsize$\pm$1.37} & 98.63{\scriptsize$\pm$0.03} & 97.44{\scriptsize$\pm$0.03} & \textbf{95.73{\scriptsize$\pm$0.32}} & 94.61{\scriptsize$\pm$0.09} & 57.74{\scriptsize$\pm$3.55} & 68.94{\scriptsize$\pm$1.18} \\
 & SegFormer & 78.22{\scriptsize$\pm$0.15} & 85.97{\scriptsize$\pm$0.12} & 97.28{\scriptsize$\pm$0.07} & 94.42{\scriptsize$\pm$0.62} & 91.16{\scriptsize$\pm$0.68} & 90.12{\scriptsize$\pm$0.37} & 44.96{\scriptsize$\pm$1.93} & 51.42{\scriptsize$\pm$1.48} & 98.62{\scriptsize$\pm$0.03} & 97.13{\scriptsize$\pm$0.33} & 95.38{\scriptsize$\pm$0.37} & 94.80{\scriptsize$\pm$0.20} & 62.01{\scriptsize$\pm$1.82} & 67.91{\scriptsize$\pm$1.28} \\
 & UPerNet$_{\text{EN-B0}}$ & 79.39{\scriptsize$\pm$0.63} & 87.00{\scriptsize$\pm$0.56} & 97.45{\scriptsize$\pm$0.02} & 94.48{\scriptsize$\pm$0.24} & 91.09{\scriptsize$\pm$0.40} & 90.11{\scriptsize$\pm$0.27} & 48.60{\scriptsize$\pm$2.60} & \textbf{54.64{\scriptsize$\pm$2.57}} & 98.71{\scriptsize$\pm$0.01} & 97.16{\scriptsize$\pm$0.13} & 95.34{\scriptsize$\pm$0.22} & 94.80{\scriptsize$\pm$0.15} & 65.38{\scriptsize$\pm$2.38} & \textbf{70.64{\scriptsize$\pm$2.13}} \\
 & UPerNet$_{\text{MiT-B0}}$ & \textbf{79.66{\scriptsize$\pm$0.46}} & \textbf{87.17{\scriptsize$\pm$0.40}} & \textbf{97.49{\scriptsize$\pm$0.11}} & \textbf{95.18{\scriptsize$\pm$0.23}} & 91.51{\scriptsize$\pm$0.51} & \textbf{90.50{\scriptsize$\pm$0.43}} & \textbf{49.68{\scriptsize$\pm$1.76}} & 53.62{\scriptsize$\pm$2.76} & \textbf{98.73{\scriptsize$\pm$0.05}} & \textbf{97.53{\scriptsize$\pm$0.12}} & 95.57{\scriptsize$\pm$0.28} & \textbf{95.01{\scriptsize$\pm$0.24}} & \textbf{66.37{\scriptsize$\pm$1.58}} & 69.78{\scriptsize$\pm$2.33} \\

\bottomrule
\end{tabular}%
}
\label{tab:results_6class}
\vspace{-2ex}
\end{table*}

\vspace{-0.5ex}
\subsection{Effect of Spatial Mapping: RGB$_{\text{ori}}$ vs. RGB$_{\text{reg}}$}
\vspace{-1ex}
\label{sec:resolution_cost}
Mapping RGB$_{\text{ori}}$ (704$\times$1280) to the registered NIR resolution (192$\times$384) induces a near 12x difference in the number of pixels, resulting in a performance drop of RGB$_{\text{reg}}$ for all classes across the evaluated SSMs, as found in Table~\ref{tab:results_6class}. The reductions in mIoU points are: 4.44 for UNet, 4.01 for UNet-CBAM, 4.72 for DeepLabV3+, 4.67 for SegFormer, 5.97 for UPerNet$_{\text{EN-B0}}$, and 5.70 for UPerNet$_{\text{MiT-B0}}$, for an average drop of 4.92 points. The UPerNet variants exhibit the largest absolute reductions despite achieving the best performance. 

The effect is strongly class-dependent, averaged across SSMs: Water drops IoU by 12.68 points and Grass by 8.89 points, compared with 2.95 for Conc., 2.34 for Asphalt, 2.13 for Dirt, and only 0.51 for Bkg. Water and Grass are therefore the classes most affected by the changes in spatial resolution and applied registration.
While this work establishes a strong baseline across modalities, isolating the effects of spatial resolution or boundary complexity remains for future work. Nonetheless, the newly aligned RGB-NIR offers a valuable modality for future multimodal research on HSI-Road.

\vspace{-1.25ex}
\subsection{Matched resolution: NIR vs. RGB$_{\text{reg}}$ vs. RGBN$_{\text{stk}}$}
\vspace{-1ex}
\label{sec:results_Fusion_Value}
\textbf{NIR vs. RGB$_{\text{reg}}$}: At the matched 192$\times$384 spatial resolution, NIR outperforms RGB$_{\text{reg}}$ in two SSMs (SegFormer by 0.94 mIoU points and UPerNet$_{\text{EN-B0}}$ by 1.09), but decreases performance in four (UNet: 2.03, UNet-CBAM: 2.06, DeepLabV3+: 1.42, and UPerNet$_{\text{MiT-B0}}$: 3.61). The UPerNet$_{\text{MiT-B0}}$ on NIR also exhibits the lowest consistency at $73.42\pm3.11$\% mIoU, indicating greater run-to-run variability than the other evaluated configurations.

\textbf{RGBN$_{\text{stk}}$ vs. RGB$_{\text{reg}}$}: RGBN$_{\text{stk}}$ achieves higher mean mIoU than RGB$_{\text{reg}}$ for five of the six SSMs (UNet: 1.44, DeepLabV3+: 1.24, SegFormer: 3.52, UPerNet$_{\text{EN-B0}}$: 3.11, and UPerNet$_{\text{MiT-B0}}$: 2.63), whereas UNet-CBAM decreases slightly by 0.14 points. For the five SSMs, RGBN$_{\text{stk}}$ narrows the mIoU gap between RGB$_{\text{ori}}$ and RGB$_{\text{reg}}$. 

\vspace{-2ex}
\textbf{Class-level Performance}: The largest consistently positive class-level gains are observed for Water, for which RGBN$_{\text{stk}}$ exceeds RGB$_{\text{reg}}$ by 4.02--10.20 points for all six SSMs. Asphalt and Dirt also show higher IoU for all six SSMs, but with smaller gains, while Conc. improves for five of six SSMs.~Grass improves for four SSMs and decreases for UNet and UNet-CBAM, indicating a relatively less consistent benefit. Among the three matched-resolution modalities, RGBN$_{\text{stk}}$ achieves the highest performance, although the best-performing SSM--modality configuration varies by class. This result supports the potential value of complementary RGB--NIR information.

\begin{table}[htbp]
\vspace{-2ex}
  \centering
  \small
  \caption{Model complexity and forward-pass latencies.}
  \vspace{-1ex}
  \label{tab:inference_performance}
  \resizebox{\columnwidth}{!}{%
  \begin{tabular}{p{1cm}l r r r r}
    \toprule
    \textbf{Modality} & \textbf{ SSMs} & \textbf{Params (M)} & \multicolumn{2}{c}{\textbf{Mean Latency *}} & \textbf{Peak GPU} \\
    \cmidrule(lr){4-5}
    & & & {\textbf{CPU (s)}} & {\textbf{GPU (ms)}} & {\textbf{Mem (MiB)}} \\
    \midrule
    \multirow{5}{*}{RGB$_{\text{ori}}$} 
      & UNet                 & 6.25  & 4.78 & 12.50 &  445.65 \\
      & DeepLabV3+           & 4.91  & 4.77 & 10.56 &  337.95 \\
      & SegFormer            & 3.72  & 3.63 & 10.02 &  695.44 \\
      & UPerNet$_{\text{MiT-B0}}$      & 10.73 & 4.27 & 14.24 & 1002.60 \\
      & UPerNet$_{\text{EN-B0}}$          & 11.61 & 4.28 & 14.35 & 1003.13 \\
    \midrule
    \multirow{5}{*}{RGB$_{\text{reg}}$} 
      & UNet                 & 6.25  & 4.43 & 11.47 &  91.53 \\
      & DeepLabV3+           & 4.91  & 4.49 & 10.26 &  86.93 \\
      & SegFormer            & 3.72  & 2.95 &  9.22 &  99.89 \\
      & UPerNet$_{\text{MiT-B0}}$      & 10.73 & 3.47 & 10.36 & 166.94 \\
      & UPerNet$_{\text{EN-B0}}$          & 11.61 & 3.46 & 10.30 & 166.94 \\
    \midrule
    \multirow{5}{*}{NIR} 
      & UNet                 & 6.26  & 4.39 & 11.03 &  98.58 \\
      & DeepLabV3+           & 4.92  & 4.54 & 10.69 &  93.99 \\
      & SegFormer            & 3.75  & 3.02 &  8.69 & 106.21 \\
      & UPerNet$_{\text{MiT-B0}}$      & 10.77 & 3.46 & 10.50 & 173.54 \\
      & UPerNet$_{\text{EN-B0}}$          & 11.62 & 3.48 & 10.35 & 173.54 \\
    \midrule
    \multirow{5}{*}{RGBN$_{\text{stk}}$} 
      & UNet                 & 6.26  & 4.36 & 11.04 &  99.27 \\
      & DeepLabV3+           & 4.92  & 4.49 & 10.27 &  94.58 \\
      & SegFormer            & 3.75  & 2.96 &  8.75 & 107.63 \\
      & UPerNet$_{\text{MiT-B0}}$      & 10.77 & 3.47 & 10.18 & 176.03 \\
      & UPerNet$_{\text{EN-B0}}$          & 11.62 & 3.48 & 10.33 & 176.15 \\
    \bottomrule
  \end{tabular}
  }
  \parbox{\columnwidth}{\raggedright\fontsize{7.25pt}{9pt}\selectfont
* Averaged over test set (570 images) with batch size 1, after 10 warmup iterations.\\
}
\vspace{-4ex}
\end{table}

\vspace{-2ex}
\subsection{Backbone Comparison}
\vspace{-1ex}
\textbf{Overall Performance}: UPerNet variants achieve the highest mIoU and mF1 across input modalities: UPerNet$_{\text{MiT-B0}}$ leads for RGB$_{\text{ori}}$, RGB$_{\text{reg}}$, and RGBN$_{\text{stk}}$, whereas UPerNet$_{\text{EN-B0}}$ performs best for NIR. Among non-UPerNet SSMs, SegFormer with RGBN$_{\text{stk}}$ achieves the strongest result, reaching 78.22\% mIoU and 85.97\% mF1, and also exhibits the largest gain, improving over RGB$_{\text{reg}}$ by 3.52 mIoU and 2.92 mF1 points. The gap between the two UPerNet variants, 3.95 mIoU and 3.45 mF1 points, on NIR highlights the importance of encoder choice even when the decoder is fixed.

\textbf{Spatial-Mapping Sensitivity}: All SSMs exhibit reduced performance when RGB$_{\text{ori}}$ is mapped to RGB$_{\text{reg}}$, with decreases of 4.01--5.97 mIoU and 3.57--4.80 mF1 points. UNet-CBAM exhibits the smallest reduction (4.01 mIoU and 3.57 mF1 points ). For RGB$_{\text{reg}}$, UPerNet$_{\text{MiT-B0}}$ combines the highest mean performance with the lowest cross-seed variability (77.03$\pm$0.27 mIoU and 84.99$\pm$0.24 mF1). At RGB$_{\text{ori}}$, it achieves the highest mean performance, although UNet exhibits lower cross-seed variability.

\vspace{-2ex}
\subsection{Computational Efficiency}
\vspace{-1ex}
Table~\ref{tab:inference_performance} shows forward-pass latency and computational requirements on the evaluated hardware (Table~\ref{tab:training_hyperparameters}) for each~SSM--modality configuration. SegFormer has the fewest parameters (3.72--3.75M) and the lowest CPU and GPU latency across all input modalities, including 2.96s on CPU and 8.75ms on GPU for RGBN$_{\text{stk}}$.~The UPerNet variants use 10.73--11.62M parameters, 2.87--3.12$\times$ the parameters in SegFormer, while achieving higher six-class segmentation accuracy.~For RGBN$_{\text{stk}}$, the UPerNet variants require 10.18--10.33ms GPU latency, compared with 8.75ms for SegFormer. DeepLabV3+ has the lowest peak GPU-memory consumption across all inputs, whereas SegFormer has the lowest parameter count and inference latency. Moreover, RGBN$_{\text{stk}}$ reduces peak GPU memory by 72.0\% (DeepLabV3+) up to 84.5\% (SegFormer) compared to RGB$_{\text{ori}}$. The results therefore demonstrate a trade-off between segmentation accuracy, latency, parameter count, and memory requirements.


\vspace{-1.25ex}
\section{Limitations and Future Work}
\vspace{-1ex}
\label{sec:limitations}

\textbf{Training from scratch.} All SSMs are trained from scratch in this work to maintain a consistent comparison across input modalities. Future work should investigate HSI-specific pretrained vision and hyperspectral foundation models.


\textbf{RGB--NIR stacking.} RGBN$_{\text{stk}}$ uses channel stacking: Intermediate, late, and attention-based fusion approaches may exploit cross-modal structure more effectively. Band selection and dimensionality reduction should also be investigated due to HSI's inherent adjacent-band redundancies.

\textbf{Registration and resolution are confounded.}
The current work does not isolate the individual contributions of spatial resolution (Table~\ref{tab:training_hyperparameters}), class distribution (Table~\ref{tab:dataset_class_distribution}), boundary complexity, or data splitting strategy (Table~\ref{tab:stratified_class_distribution}). Isolating these confounding factors remains a subject for future work.

\vspace{-1.25ex}
\section{Conclusion}
\vspace{-1ex}
\label{sec:conclusion}
This work extends the HSI-Road dataset with (i) a manually annotated six-class surface-oriented taxonomy, (ii) a pipeline for registering RGB images and corresponding relabeled masks to the NIR resolution, and (iii) a reference multilabel-stratified train/validation/test split.
Experiments show that segmentation performance on the original binary masks is tightly clustered, whereas the six-class taxonomy reveals clearer differences among input modalities and SSMs.

RGB$_{\text{ori}}$ produces the highest overall result, both at the SSM level and per class, with UPerNet$_{\text{MiT-B0}}$ reaching 82.73\% mIoU and 89.43\% mF1. Mapping RGB$_{\text{ori}}$ to RGB$_{\text{reg}}$ reduces mIoU by 4.01--5.97 points and mF1 by 3.57--4.80 points across all SSMs. However, RGB$_{\text{ori}}$ (704$\times$1280) has 12$\times$ the pixel count of RGB$_{\text{reg}}$ and RGBN$_{\text{stk}}$ requires 3.57--6.46$\times$ less peak GPU memory than RGB$_{\text{ori}}$. RGB$_{\text{ori}}$ should therefore be regarded as a high-resolution reference rather than a matched-resolution test of RGB against other input modalities.

At the matched 192$\times$384 resolution, RGBN$_{\text{stk}}$ improves over NIR for all six SSMs and over RGB$_{\text{reg}}$ for five of six SSMs, with gains of up to 3.52 mIoU and 2.92 mF1 points over RGB$_{\text{reg}}$. Water exhibits the most consistent class-level improvement, whereas the effect on Grass remains SSM-dependent.
Among the evaluated SSMs, the UPerNet variants achieve the highest mean segmentation performance, while SegFormer offers the lowest parameter count and measured inference latency, alongside the largest RGBN$_{\text{stk}}$ gain. 
 

These results suggest three practical considerations for surface-oriented road-scene segmentation: (i) modality comparisons should be made at matched resolution, since spatial resolution has a substantial effect on performance; (ii) once resolution is matched, complementary RGB–NIR information can yield a measurable but SSM- and class-dependent benefit; and (iii) SSM--Encoder choice should be driven by deployment constraints as much as by segmentation accuracy.

\vspace{-1.25ex}
\let\oldthebibliography\thebibliography
\let\endoldthebibliography\endthebibliography
\renewenvironment{thebibliography}[1]{%
  \begin{oldthebibliography}{#1}%
    \setlength{\itemsep}{0pt}
    \setlength{\parsep}{0pt}
    \setlength{\baselineskip}{11.5pt}
}{%
  \end{oldthebibliography}%
}

\bibliographystyle{IEEEtran}
\bibliography{refs}

\clearpage
\onecolumn

\end{document}